\documentclass[letterpaper]{article}
\usepackage[preprint]{aaai2027}
\usepackage{etoolbox}
\makeatletter
\patchcmd{\@maketitle}
  {\Large{\textbf{\aaai@theauthors\ifhmode\\\fi}}}
  {{\normalsize\bfseries\aaai@theauthors\ifhmode\\\fi}}
  {\typeout{Patched author size (1)}}{\typeout{FAILED author size (1)}}
\patchcmd{\@maketitle}
  {\Large{\textbf{\aaai@theauthors\ifhmode\\\fi}}}
  {{\normalsize\bfseries\aaai@theauthors\ifhmode\\\fi}}
  {\typeout{Patched author size (2)}}{\typeout{FAILED author size (2)}}
\patchcmd{\@maketitle}
  {\vskip 1em plus 2fil}
  {\vskip 2.25em plus 2fil}
  {\typeout{Patched title gap (1)}}{\typeout{FAILED title gap (1)}}
\patchcmd{\@maketitle}
  {\vskip 1em plus 2fil}
  {\vskip 2.25em plus 2fil}
  {\typeout{Patched title gap (2)}}{\typeout{FAILED title gap (2)}}
\makeatother
\usepackage[hyphens]{url}
\usepackage{graphicx}
\usepackage{natbib}
\usepackage{caption}
\usepackage{amsmath}
\usepackage{amssymb}
\usepackage{booktabs}
\usepackage{multirow}
\usepackage{array}
\usepackage{enumitem}
\usepackage{dblfloatfix}
\usepackage{tcolorbox}

\newtcolorbox{teaserbox}{
    colback=gray!12,
    colframe=white,
    boxsep=6pt,
    left=8pt, right=8pt, top=6pt, bottom=6pt
}

\title{Speaking the Navigator's Language: Trajectory-Grounded Instruction Translation for Frozen Aerial VLN Agents}
\author{
    Xi Chen\textsuperscript{\rm 1},
    Zhe Liu\textsuperscript{\rm 1},
    Xiaogang Xu\textsuperscript{\rm 1},
    Jiafei Wu\textsuperscript{\rm 1},
    Chunyi Zhou\textsuperscript{\rm 1},
    Yuan Su\textsuperscript{\rm 1},
    Rui Zeng\textsuperscript{\rm 1},
    Tianyu Du\textsuperscript{\rm 1},\\
    Kelu Yao\textsuperscript{\rm 2},
    Chao Li\textsuperscript{\rm 2},
    Shouling Ji\textsuperscript{\rm 1}
}
\affiliations{
    \textsuperscript{\rm 1}Zhejiang University\\
    \textsuperscript{\rm 2}Zhejiang Lab\\
    \{chan\_xi, zhe.liu\}@zju.edu.cn,
    xiaogangxu00@gmail.com,\\
    \{jiafeiwu, zhouchunyi, yuansu, ruizeng24, zjradty\}@zju.edu.cn,\\
    \{yaokelu, lichao\}@zhejianglab.org,
    sji@zju.edu.cn
}

\begin{document}
\maketitle

\begin{abstract}
Aerial vision-and-language navigation (VLN) agents are typically trained on detail-rich, trajectory-aligned commands, whereas users issue short, intent-driven instructions; on a frozen OpenFly navigator, this \emph{instruction gap} drops success rate (SR) from $31.03\%$ to $11.33\%$.
To scale translator training, we prompt a language model with human-written style examples to convert original commands into paired, intent-centered Weak commands, which yield $15.27\%$ SR.
We introduce the \textbf{Trajectory-Grounded Instruction Translator (TGIT)}, a front-end that keeps the navigator frozen and translates Weak inputs into agent-executable commands by learning from its trajectory outcomes.
The resulting Weak-trained translator raises Weak-input SR to $37.93\%$ and transfers zero-shot to real human instructions ($11.33\%{\rightarrow}32.51\%$); it also improves held-out OpenFly ($4.95\%{\rightarrow}20.79\%$) and yields recovery on CityNav and AirVLN.
\end{abstract}

\section{Introduction}

Vision-and-Language Navigation (VLN) has progressed from indoor instruction following~\cite{anderson2018vision,hao2020towards,qiao2022hop,zhang2024vision} to aerial platforms, where UAV agents must act under continuous 3D motion, large viewpoint changes, and fine-grained spatial constraints~\cite{liu2023aerialvln,yao2024aeroverse,wang2024towards,saxena2025uav,sautenkov2025uav}.
Yet aerial VLN agents are commonly trained on detail-dense, trajectory-aligned instructions produced by templates or vision--language generators~\cite{gao2025openfly,lee2024citynav,wang2023scaling,fan2024navigation}, whereas users tend to issue shorter, intent-driven commands.

\begin{figure}[t]
  \centering
  \includegraphics[width=\linewidth]{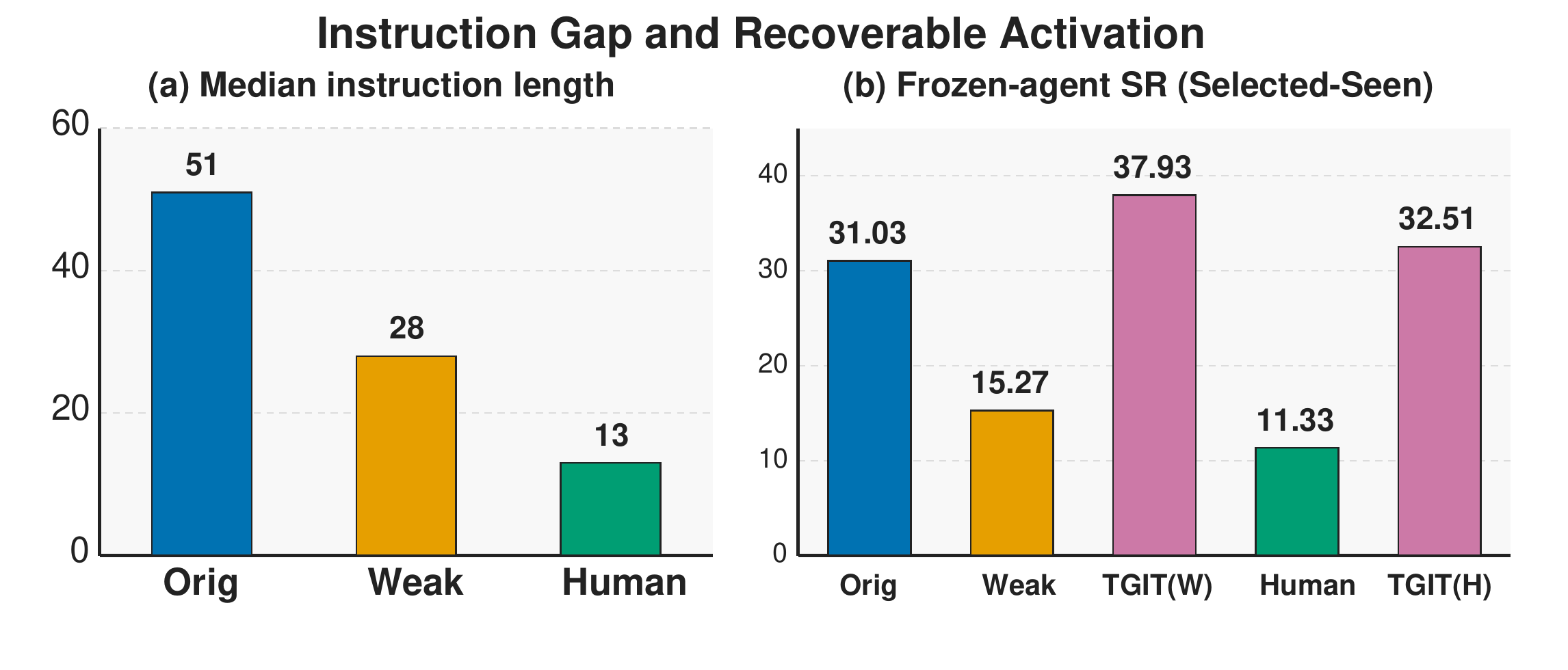}
  \caption{\textbf{Instruction gap and recoverable activation on a frozen OpenFly-Agent.}
  (a) Median instruction length collapses from Orig to Weak to Human.
  (b) On the 203-episode evaluation set (Selected-Seen), SR drops from $31.03\%$ (Orig) to $15.27\%$ (Weak) and $11.33\%$ (Human). TGIT(W) recovers to $37.93\%$ on Weak; the same adapter transfers zero-shot as TGIT(H) to $32.51\%$ on Human.}
  \label{fig:intro_gap}
\end{figure}

We begin with a simple controlled probe: keeping the routes, initial conditions, and UAV navigator fixed, we change only the instruction source.
If the navigator understood its training and deployment interfaces equally well, its performance should remain stable.
Instead, using natural instructions collected from 16 annotators for 203 OpenFly episodes, SR drops from $31.03\%$ with dataset instructions to $11.33\%$ with Human instructions (Figure~\ref{fig:intro_gap}).
This behavioral collapse reveals an \emph{instruction gap}: deployment language fails to activate capabilities available through the same frozen agent's training-language interface.

\begin{center}
\begin{teaserbox}
\small
\textbf{Example: Instruction Gap in UAV Navigation.}
\label{ex:uav_gap}

\medskip
\textbf{Dataset instruction:} \\
``Turn right and proceed straight for 30 meters until reaching the gray rectangular building with a flat roof.''

\medskip
\textbf{Human-style instruction:} \\
``Fly toward the building ahead.''
\end{teaserbox}
\end{center}

Closing this gap is hard for a frozen navigator trained on trajectory-aligned Orig commands: short Human instructions must be rewritten into language that matches its learned associations, and two equally plausible expansions can still induce different actions.
Agent-centric retraining methods can absorb this mismatch~\cite{kamath2023new,wang2020vision}, but are costly and may change perception and control, making it hard to tell whether gains come from adapting to new language or from altering the navigator itself~\cite{gao2025openfly,lee2024citynav,liu2023aerialvln,wang2024towards}.
Instruction-centric rewriting methods instead map weak inputs back to an Orig sentence~\cite{shah2023lm,zhou2024navgpt,zhang2023vlntrans}, yet matching that text does not tell whether a particular navigator will succeed.
We therefore keep the navigator fixed and learn agent-executable translations from its trajectory outcomes, not from text similarity alone.

This observation also determines which failures should supervise the translator.
Training on arbitrary failures would conflate an instruction-interface mismatch with navigation skills that the backbone does not possess.
We therefore focus on paired cases where the original instruction succeeds, but its weakened counterpart fails under the same route, initial condition, simulator, and policy.
Orig success serves as a capability witness, while Weak failure isolates a recoverable interface intervention.
Human annotation is too costly to scale for translator training, so we construct Weak instead: using a few Human-written instructions as style examples, an LM converts each Orig command into a shorter, intent-centered counterpart.

Building on these choices, we introduce \textbf{TGIT} (\textbf{T}rajectory-\textbf{G}rounded \textbf{I}nstruction \textbf{T}ranslator).
It freezes the UAV policy and adapts only a LoRA VLM front-end that translates Weak inputs into agent-executable commands.
Starting from Orig instructions in the target agent's training dataset and their LM-generated Weak counterparts, TGIT selects hard cases, proposes multiple candidates, executes them with the frozen navigator, and converts weighted trajectory outcomes into preferences for multi-round DPO with pool resampling.
The resulting translator is then evaluated on Weak and transferred zero-shot to Human, testing whether execution-grounded preferences activate the agent without imitating a single Orig sentence.

On the 203-episode OpenFly evaluation, TGIT raises SR from $15.27\%$ to $37.93\%$ ($2.48{\times}$), exceeding Orig ($31.03\%$), ICL, and SFT; the same Weak-trained translator raises Human-input SR zero-shot from $11.33\%$ to $32.51\%$ ($2.87{\times}$).
Held-out OpenFly scenes and two additional frozen agents show the same relative activation, but their absolute success remains bounded by the navigator itself.
These results support a bounded claim: TGIT repairs recoverable instruction-interface failures, while strengthening the frozen navigator remains an orthogonal agent-centric problem.

\noindent\textbf{Our contributions are threefold:}
\begin{itemize}
    \item \textbf{Recoverable instruction gap.}
    We provide Human evidence for the mismatch between deployment and training language, and operationally identify recoverable failures through original-success / Human-failure pairs.
    \item \textbf{Agent-executable instruction translation.}
    We introduce TGIT, a plug-and-play front-end that can be applied to any frozen VLN agent using only that agent's own training instructions and their Weak counterparts: it builds trajectory-grounded preference data from hard cases and multi-round DPO to translate underspecified inputs into executable commands.
    \item \textbf{Controlled cross-agent validation.}
    TGIT outperforms Weak, ICL, and SFT on OpenFly, transfers zero-shot from Weak training to Human inputs, and yields consistent relative gains across held-out environments and multiple frozen UAV agents.
\end{itemize}
The appendix complements the main results with preliminary evidence, validity checks, and reproducibility details.
A preliminary study further characterizes training-instruction style and agent sensitivity to underspecification, strengthening the instruction-gap motivation.
Paired lexical and embedding analyses further show that Weak moves from Orig toward Human.
Complete prompts, dataset statistics, optimization settings, translator-scale diagnostics, and same-route trajectory comparisons support reproduction and expose how the learned translations change agent behavior.
\section{Related Work}

Prior VLN and embodied-language research largely falls into two axes: improving the navigator, and rewriting or generating language for clarity and coverage~\cite{zhang2024vision,lin2023advances,ma2024survey}.
TGIT is not another point on either axis.
It keeps the aerial policy frozen and learns an instruction interface that activates competence already present in that policy.

\paragraph{Agent-centric aerial VLN.}
VLN was established in photorealistic indoor environments~\cite{anderson2018vision} and later extended toward continuous control, long-horizon navigation, and sim-to-real evaluation~\cite{anderson2021sim,song2025towards,lin2025vlnverse}.
AerialVLN introduced continuous outdoor UAV flight, while CityNav, OpenFly, and AeroVerse scale aerial environments, trajectories, and embodied controllers~\cite{liu2023aerialvln,lee2024citynav,gao2025openfly,yao2024aeroverse}.
These advances improve navigation capability, typically under detailed, trajectory-aligned training instructions.

\paragraph{Instruction-centric language interfaces.}
Synthetic instruction generation scales VLN supervision~\cite{huang2022assister,kamath2023new,wang2023scaling,gopinathan2024spatially}, while landmark-centered generation and ULN expose the importance of specificity~\cite{wang2022less,feng2022uln}.
Closest to our setting, VLN-Trans decomposes instructions into landmark-discriminative sub-instructions~\cite{zhang2023vlntrans}; T-araVLN translates language for agricultural navigation~\cite{zhao2025taravln}; and RAM rewrites observation--instruction pairs for augmentation and unseen generalization~\cite{wei2025ram}.
These methods optimize clarity, recognizability, or training-data coverage~\cite{wang2023lana}.

\paragraph{Our position.}
TGIT never updates the UAV navigator, and it ranks candidate translations with simulator rollouts and updates only a LoRA front-end via preference learning~\cite{rafailov2023direct,an2023direct,calandriello2024human}.
Multimodal preference methods often align VLM responses or retrain the VLA itself~\cite{xing2025realign,hung2025nora}; TGIT instead uses original-success / weak-failure hard cases to restrict supervision to episodes where the navigator has already demonstrated the required skill.
Thus TGIT targets interface activation rather than policy retraining, instruction imitation, or data augmentation.

\begin{figure*}[t]
  \centering
  \includegraphics[width=0.98\linewidth]{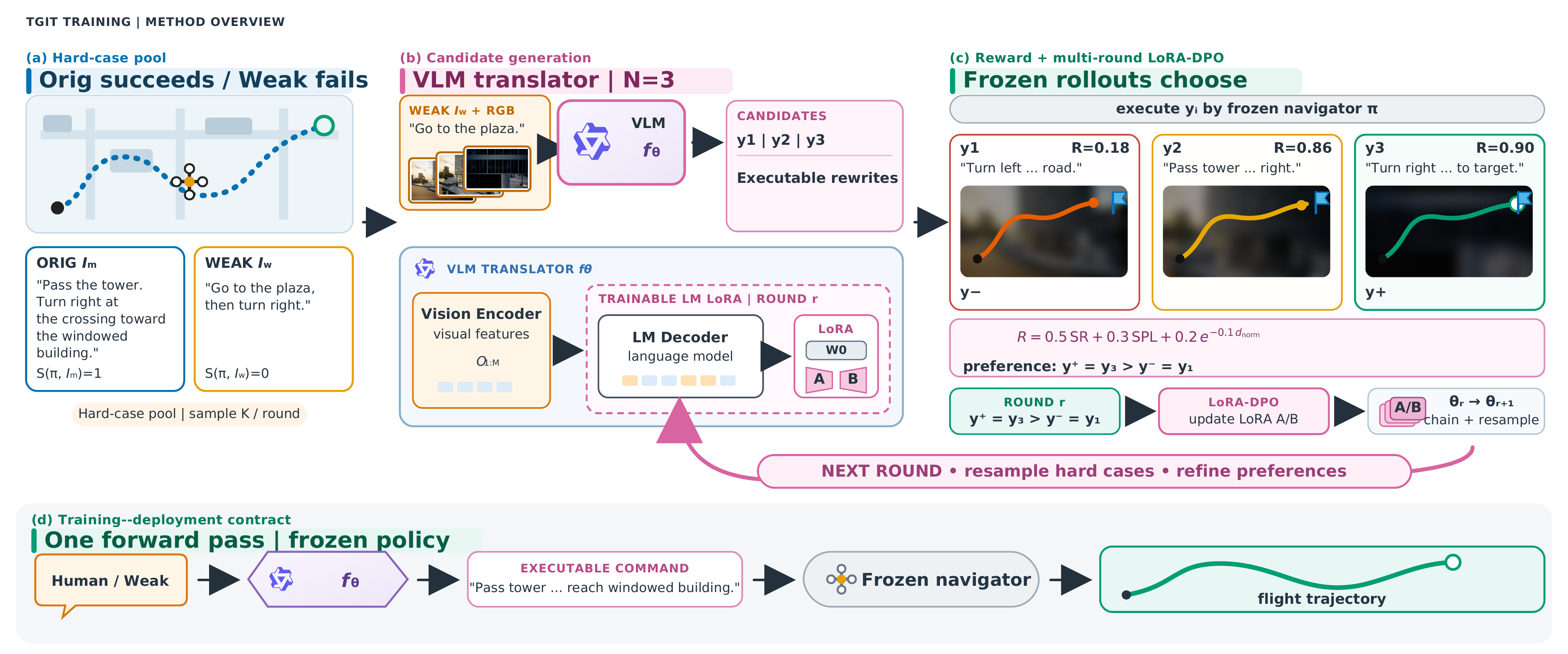}
  \caption{
\textbf{The overview of TGIT.}
(a)~\emph{Hard-case pool}: keep episodes with $S(\pi,I_m){=}1$ and $S(\pi,I_w){=}0$ so that capability exists but the weak interface fails.
(b)~\emph{Candidate generation}: VLM translator $f_\theta$ takes $(I_w,\mathcal{O}_{1:M})$ and emits $N{=}3$ executable rewrites; only LoRA $A$/$B$ on the LM decoder is trainable.
(c)~\emph{Reward$+$multi-round LoRA--DPO}: frozen rollouts score each $y_i$ by $R{=}0.5\,\mathrm{SR}{+}0.3\,\mathrm{SPL}{+}0.2\,e^{-0.1 d_{\mathrm{norm}}}$ (SR: success rate; SPL: path efficiency; $d_{\mathrm{norm}}$: normalized final distance; blue flags: targets), form preferences $y^+\!\succ\! y^-$, update $A$/$B$, then chain the adapter and resample hard cases for the next round.
(d)~\emph{Training--deployment contract}: evaluation collapses to one forward pass---no reward, preference, or policy update (Eq.~\ref{eq:objective}).
}
  \label{fig:t2rl_framework}
\end{figure*}

\section{Methodology}
\label{sec:method}

\subsection{Problem Formulation}

\paragraph{System actors and freeze boundary.}
Our system contains three actors: an aerial environment $\mathcal{E}$; a pretrained UAV navigator $\pi$; and a front-end instruction translator $f_\theta$.
The navigator consumes an instruction $I$ and executes a trajectory
$\hat{\mathcal{T}}=\pi(I,\mathcal{E}).$
Throughout adaptation and evaluation, $\pi$ and its controller are frozen; $\theta$ is the only trainable parameter set.
The translator also observes visual context $\mathcal{O}_{1:M}=\{o_m\}_{m=1}^{M}$, where $M{=}1$ is a single RGB frame and $M{>}1$ provides multi-frame context.
This follows aerial-VLN annotation protocols in which language is grounded in rendered observations~\cite{gao2025openfly,lee2024citynav,liu2023aerialvln}.
Visual context supplies scene evidence to the front-end VLM, but no rewards, actions, goal labels, or policy updates are available to it at deployment.

\paragraph{Instruction interface and objective.}
Conventional aerial VLN trains $\pi$ with detail-rich, trajectory-aligned instructions $I_m\sim\mathcal{D}_m$~\cite{gao2025openfly,lee2024citynav,wang2023scaling}, whereas deployment instructions may follow a Human distribution $\mathcal{D}_h\neq\mathcal{D}_m$.
For controlled optimization, we use Human-informed LM-weakened inputs $I_w\sim\mathcal{D}_w$; $\mathcal{D}_w$ is a paired stress distribution, not an estimator of $\mathcal{D}_h$.
Given $I_w$ and $\mathcal{O}_{1:M}$, the translator emits
$\tilde{I}=f_\theta(I_w,\mathcal{O}_{1:M}),$
which the unchanged navigator executes as
\begin{equation}
    \hat{\mathcal{T}}=\pi(\tilde{I},\mathcal{E}).
\end{equation}
Let $\mathcal{T}=(s_1,\ldots,s_T)$ denote the reference UAV trajectory, with each state containing 3D position, orientation, and altitude.
We optimize only the interface (subject to $\pi$ fixed):
\begin{equation}
\label{eq:objective}
\begin{aligned}
\max_{\theta}\quad
&\mathbb{E}_{(I_w,\mathcal{T},\mathcal{O}_{1:M})\sim\mathcal{D}_w}\\[-2pt]
&\quad
\Big[
R\big(\pi(f_\theta(I_w,\mathcal{O}_{1:M}),\mathcal{E}),\mathcal{T}\big)
\Big],
\end{aligned}
\end{equation}
where $R(\cdot)$ measures trajectory-level execution quality.

\paragraph{Design goals.}
Equation~\ref{eq:objective} induces three requirements aligned with the challenges in the Introduction.
First, \emph{capability identification}: supervision should target episodes whose required navigation skill has already been demonstrated, rather than asking translation to repair missing perception or control.
Second, \emph{frozen-interface adaptation}: the method should learn an agent-specific executable language while preserving heterogeneous UAV policies.
Third, \emph{execution-grounded supervision}: learning must use expensive, non-differentiable rollout outcomes and accommodate multiple valid command phrasings.
For these, TGIT addresses these requirements through hard-case selection, a parameter-efficient front-end, and trajectory-ranked preference optimization, respectively.

\subsection{Trajectory-Grounded Instruction Translation}
\label{sec:tgit}

\paragraph{Overview: closed training--deployment loop.}
Figure~\ref{fig:t2rl_framework} summarizes TGIT.
Training first selects original-success / weak-failure hard cases.
For each sampled weak input, $f_\theta$ generates $N$ candidate commands; the frozen navigator executes every candidate; a weighted trajectory reward converts their outcomes into pairwise preferences; and DPO updates only the LoRA translator before the hard-case pool is resampled for the next round.
At deployment, this loop collapses to one forward pass: a deployment-style instruction and available RGB frames enter $f_\theta$, one executable command is emitted, and $\pi$ flies unchanged.
Simulator rewards, preference pairs, and reference trajectories are training-only signals.
Section~\ref{sec:ablation} separately tests pool construction, reward ranking, and translator backbone choices without redefining the primary protocol.

\paragraph{Hard-case pool construction.}

Under a limited execution budget, random sampling yields many uninformative episodes: either the navigator already fails under Orig, or Weak happens to succeed by chance.
We therefore construct a compact hard-case pool $D_{\mathrm{pool}}$.
Each instance is selected such that the original dataset instruction succeeds, while the corresponding weakened human-style instruction fails under the same frozen agent.
Writing $S(\pi,I,\mathcal{E})\in\{0,1\}$ for episode success, the recoverable candidate set is
\begin{equation}
\begin{aligned}
D_{\mathrm{pool}}=\{(I_m,I_w,\mathcal{E}) :\;&
S(\pi,I_m,\mathcal{E})=1,\\[-2pt]
&
S(\pi,I_w,\mathcal{E})=0\}.
\end{aligned}
\end{equation}
These original-success / weak-failure cases isolate the instruction interface mismatch: the agent has demonstrated the required navigation ability, but the weakened command is not executable enough.
In practice, we build a pool of $100$ hard cases for each UAV agent and sample $K{=}20$ cases per round, resampling across rounds to expose diverse failure modes.
Failures under both $I_m$ and $I_w$ are not used as evidence of a recoverable language problem.
Each pair holds the route, initial condition, simulator, and policy fixed while changing only the instruction: Orig success witnesses capability, whereas Weak failure isolates the interface intervention.
Section~\ref{sec:ablation} ablates this selection and resampling against random and fixed-subset alternatives.

\paragraph{Candidate generation and execution.}
For each sampled instruction $x\in D_{\mathrm{pool}}$, the translator generates $N=3$ candidate executable commands $\{y_i\}_{i=1}^N$ conditioned on $(x,\mathcal{O}_{1:M})$ under format constraints.
The translator conditions on visual context $\mathcal{O}_{1:M}$ as an input prior; Section~\ref{sec:ablation} varies frame count (e.g., one vs.\ four frames) to test this prior.
We use complementary decoding settings: one deterministic setting for stable command structure and stochastic settings for semantic variation.
During training, the original dataset instruction is also included as a strong execution reference, but inference uses only the human-style input and available observations.
Each candidate $y_i$ is executed by the frozen UAV agent, producing a trajectory $\tau_i$ with success, path efficiency, and final-distance measurements.
Execution is essential because text similarity cannot identify an executable command: close paraphrases may induce different behavior, while distinct surface forms may produce equally successful trajectories~\cite{daniele2017navigational,liang2022code,zhou2024navgpt}.

  \paragraph{Weighted trajectory reward and preferences.}
  The primary protocol ranks candidates using
  \begin{equation}
      R(\tau_i,\mathcal{T}) =
      \alpha\,\mathrm{SR}_i + \beta\,\mathrm{SPL}_i
      + \gamma\,e^{-\lambda d_i^{\mathrm{norm}}},
  \end{equation}
  where $\mathrm{SR}_i$, $\mathrm{SPL}_i$, and $\mathrm{NE}_i$ follow standard VLN metrics~\cite{anderson2018vision}, $\mathrm{SPL}_i=\ell_i/\max(p_i,\ell_i)$ is path efficiency (zero for failures), and $d_i^{\mathrm{norm}}=\mathrm{NE}_i/\max(\ell_i,\epsilon)$ normalizes final navigation error by reference-path length.
  At evaluation time, reported SPL is the mean of these efficiencies over successful episodes only.
  We set $(\alpha,\beta,\gamma)=(0.5,0.3,0.2)$ and $\lambda=0.1$, making task success dominant while retaining dense efficiency and distance signals for failed or near-tied candidates.
  The reward-ranking ablation in Section~\ref{sec:ablation} compares Success-only and lexicographic alternatives, showing that recovery is not tied to these exact weights.
  We convert trajectory rewards into pairwise preferences:
  \begin{equation}
      y_i \succ y_j \quad \text{if} \quad R(\tau_i,\mathcal{T}) > R(\tau_j,\mathcal{T}).
  \end{equation}
  This transforms non-differentiable rollout outcomes into a stable supervision signal for preference learning.
  Unlike pure SFT on a single target phrasing, preference pairs teach the translator which of several plausible commands actually activate $\pi$.
  Near-tied outcomes are discarded to avoid supervision from simulator noise.
  The weighted scalar above remains the primary TGIT reward.

\paragraph{Multi-round DPO optimization.}
We optimize the translator with Direct Preference Optimization (DPO)~\cite{rafailov2023direct}, using LoRA~\cite{Hu2021LoRALA} to update only a small set of VLM parameters.
Each round starts from and references the round-matched SFT adapter, samples $K=20$ instructions from $D_{\mathrm{pool}}$, generates and executes candidates, constructs preference pairs with a minimum reward gap, and updates $f_\theta$.
Adapters are chained across rounds so later rounds refine an already grounded interface rather than restarting from the base VLM.
As the translator improves, its candidates produce better trajectories, which in turn provide more reliable preferences for the next round---closing the feedback loop.
Throughout this process, the UAV policy $\pi$ is never modified; all gains come from aligning human-style inputs with the frozen agent's executable command space.
Each round changes the sampled hard cases and candidates, but not the navigator or evaluation protocol.
Resampling broadens failure coverage, while adapter chaining preserves executable patterns.
Unlike general online preference collection~\cite{calandriello2024human}, TGIT evaluates language actions with a fixed policy.

\paragraph{Training--deployment contract.}
Adaptation may access simulator rollouts, reference paths, and trajectory metrics only to construct preference pairs.
Evaluation removes all three: the translator receives the deployment instruction and available RGB context, emits one command, and then exits the control loop.
The frozen navigator consumes that command using its original controller and stopping rule.
This contract prevents reward or trajectory leakage at deployment and attributes behavioral changes to the learned interface rather than to policy updates.

\section{Experiments}
\subsection{Experimental Setup}
\label{sec:exp_setting}

We evaluate TGIT on three frozen UAV navigation agents: OpenFly-Agent~\cite{gao2025openfly}, CityNav-Agent~\cite{lee2024citynav}, and AirVLN-Agent~\cite{liu2023aerialvln}.
All navigation policies remain unchanged; only the front-end translator is trained.
The translator is Qwen2.5-VL-7B-Instruct with LoRA~\cite{Hu2021LoRALA} ($r{=}4$, $\alpha{=}8$) on language modules.
For each agent, we build a hard-case pool of $100$ original-success / weak-failure instructions, sample $K{=}20$ cases per round for $5$ rounds, and generate $N{=}3$ candidates with visual frames.
DPO starts from and references the round-matched SFT adapter, uses $\beta{=}0.1$, a learning rate of $2{\times}10^{-5}$, and discards pairs with a reward gap below $0.05$.
Adapters carry across rounds; the navigator remains fixed.

These settings define the main protocol; Section~\ref{sec:ablation} ablates hard-case pool construction, reward ranking, visual frames, and translator backbone as diagnostics.

\paragraph{Splits and evaluation.}
We factor evaluation scenes by two axes used in prior aerial VLN protocol: whether the \emph{translator} was optimized on that scene (\textbf{Selected} vs.\ \textbf{Unselected}), and whether the frozen \emph{agent} trained on that scene (\textbf{Seen} vs.\ \textbf{Unseen})~\cite{gao2025openfly,lee2024citynav,liu2023aerialvln}.
Selected-Seen / Unselected-Seen / Unseen thus contain $203/202/404$ OpenFly, $326/180/220$ CityNav, and $120/192/147$ AirVLN episodes.
Selected-Seen is the pool-construction scene for TGIT; Unselected-Seen is agent-seen but translator-unselected; Unseen is held out for the agent.
Each closed-loop rollout lasts at most $100$ agent steps, and success requires final distance below $20$\,m.

\paragraph{Baselines and Human transfer.}
We compare Orig, Weak, ICL, and SFT under the Weak protocol; TGIT(W) is the primary result.
On OpenFly Selected-Seen, sixteen annotators provide one natural instruction per episode.
Human anchors the deployment gap; Weak is the scalable, Human-informed stress protocol.
We also report zero-shot TGIT(H), which only reuses the TGIT(W) adapter.

\paragraph{Protocol consistency.}
Within each agent and split, methods use the same episode set, initial conditions, rollout horizon, success threshold, controller, and metric implementation.
Only the instruction path and translator update differ.
Table~\ref{tab:main_results} reports the complete matched protocol; reward and translator-backbone ablations are explicitly diagnostic because they reuse narrower candidate or frame settings.

\begin{center}
\small
\noindent\textbf{Controlled comparison.}
All methods keep the navigator frozen; Human is gap evidence rather than an adaptation baseline.
\par\smallskip
\centering
\resizebox{\linewidth}{!}{
\begin{tabular}{lcccc}
\toprule
\textbf{Method} & \textbf{Eval input} & \textbf{Translator update} & \textbf{Rollout preference} & \textbf{Multi-round} \\
\midrule
Orig  & Dataset & None & No & No \\
Weak  & Weak    & None & No & No \\
Human & Human   & None & No & No \\
TGIT(H) & Human & Reuse TGIT(W) adapter & No & No \\
ICL   & Weak    & Prompt only & No & No \\
SFT   & Weak    & Supervised LoRA & No & No \\
TGIT(W) & Weak & DPO LoRA & Yes & Yes \\
\bottomrule
\end{tabular}
}
\end{center}

We report Success Rate (SR), success-conditional path efficiency (SPL), and Navigation Error (NE)~\cite{anderson2018vision,gao2025openfly,liu2023aerialvln}.
SR is the fraction of episodes that reach the goal.
SPL is the mean of $\ell/\max(p,\ell)$ over successful episodes only; failed episodes are excluded.
Thus SPL measures how efficiently successful trajectories reach the goal and is not upper-bounded by SR.
Higher SR/SPL and lower NE are better.

\begin{table*}[!t]
    \centering
    \caption{
    Main navigation results on three frozen UAV agents.
    We report Success Rate (SR, \%), success-conditional path efficiency (SPL, \%), and Navigation Error (NE, meters).
    SPL is averaged only over successful episodes.
    TGIT(H) and TGIT(W) denote Human and Weak evaluation inputs, respectively.
    \textbf{Human} and \textbf{TGIT(H)} are evaluated only on OpenFly Selected-Seen (203 episodes) due to the high cost of human annotation; other Human-input cells are ``--''.
    TGIT(H) reuses the Weak-trained TGIT(W) adapter zero-shot, without Human-specific training.
    TGIT(W) deltas are against Orig (negative NE is better).
    }
    \label{tab:main_results}
    \resizebox{0.95\textwidth}{!}{
    \begin{tabular}{@{}l|ccc|ccc|ccc@{}}
    \toprule
    \multirow{3}{*}{Method}
    & \multicolumn{9}{c}{\textbf{OpenFly-Agent} (\citeauthor{gao2025openfly},\citeyear{gao2025openfly})} \\
    \cmidrule(lr){2-10}
    & \multicolumn{3}{c}{Selected-Seen}
    & \multicolumn{3}{c}{Unselected-Seen}
    & \multicolumn{3}{c}{Unseen} \\
    & SR$\uparrow$ & SPL$\uparrow$ & NE$\downarrow$
    & SR$\uparrow$ & SPL$\uparrow$ & NE$\downarrow$
    & SR$\uparrow$ & SPL$\uparrow$ & NE$\downarrow$ \\
    \midrule
    Orig   &31.03  &31.66  &116.56  &20.05  &16.82  &77.42   &11.39  &29.82  &172.77 \\
    Weak   &15.27  &37.36  &82.48   &14.11  &13.47  &202.27  &4.95   &8.71   &90.25  \\
    Human  &11.33  &89.43  &108.92  &--     &--     &--      &--     &--     &-- \\
    \textbf{TGIT(H)}
           &\textbf{32.51}  &\textbf{92.75}  &\textbf{95.31}   &--     &--     &--      &--     &--     &-- \\
    ICL    &19.70  &18.50  &72.08   &14.60  &9.01   &145.81  &4.46   &29.74  &138.04 \\
    SFT    &29.55  &33.49  &101.79  &19.55  &14.48  &112.46  &10.40  &26.90  &98.69  \\
    \textbf{TGIT(W)}
           & \textbf{37.93}{\scriptsize(+6.90)} & \textbf{24.85}{\scriptsize(-6.81)} & \textbf{72.08}{\scriptsize(-44.48)}
           & \textbf{25.25}{\scriptsize(+5.20)} & \textbf{12.02}{\scriptsize(-4.80)} & \textbf{118.82}{\scriptsize(+41.40)}
           & \textbf{20.79}{\scriptsize(+9.40)} & \textbf{29.91}{\scriptsize(+0.09)} & \textbf{76.12}{\scriptsize(-96.65)} \\
    \midrule\midrule
    \multirow{3}{*}{Method}
    & \multicolumn{9}{c}{\textbf{CityNav-Agent} (\citeauthor{lee2024citynav},\citeyear{lee2024citynav})} \\
    \cmidrule(lr){2-10}
    & \multicolumn{3}{c}{Selected-Seen}
    & \multicolumn{3}{c}{Unselected-Seen}
    & \multicolumn{3}{c}{Unseen} \\
    & SR$\uparrow$ & SPL$\uparrow$ & NE$\downarrow$
    & SR$\uparrow$ & SPL$\uparrow$ & NE$\downarrow$
    & SR$\uparrow$ & SPL$\uparrow$ & NE$\downarrow$ \\
    \midrule
    Orig   &8.89  &90.51  &73.15   &3.68  &98.62  &95.00   &2.73  &98.30  &92.81  \\
    Weak   &6.11  &98.21  &120.60  &1.84  &98.55  &99.43   &0.00  &0.00   &93.86  \\
    ICL    &8.83  &90.51  &68.72   &1.53  &89.20  &100.28  &0.00  &0.00   &92.43  \\
    SFT    &8.59  &89.52  &80.59   &3.37  &87.23  &88.32   &2.27  &95.18  &92.28   \\
    \textbf{TGIT(W)}
           & \textbf{10.00}{\scriptsize(+1.11)} & \textbf{87.19}{\scriptsize(-3.32)} & \textbf{74.41}{\scriptsize(+1.26)}
           & \textbf{9.20}{\scriptsize(+5.52)} & \textbf{90.05}{\scriptsize(-8.57)} & \textbf{70.01}{\scriptsize(-24.99)}
           & \textbf{3.64}{\scriptsize(+0.91)} & \textbf{93.20}{\scriptsize(-5.10)} & \textbf{91.23}{\scriptsize(-1.58)} \\
    \midrule\midrule
    \multirow{3}{*}{Method}
    & \multicolumn{9}{c}{\textbf{AirVLN-Agent} (\citeauthor{liu2023aerialvln},\citeyear{liu2023aerialvln})} \\
    \cmidrule(lr){2-10}
    & \multicolumn{3}{c}{Selected-Seen}
    & \multicolumn{3}{c}{Unselected-Seen}
    & \multicolumn{3}{c}{Unseen} \\
    & SR$\uparrow$ & SPL$\uparrow$ & NE$\downarrow$
    & SR$\uparrow$ & SPL$\uparrow$ & NE$\downarrow$
    & SR$\uparrow$ & SPL$\uparrow$ & NE$\downarrow$ \\
    \midrule
    Orig   &5.83  &8.60  &98.23  &1.04  &3.78  &91.18  &0.00  &0.00  &127.76  \\
    Weak   &0.83  &8.40  &92.04  &0.00  &0.00  &92.04  &0.00  &0.00  &106.30  \\
    ICL    &1.64  &8.74  &94.25  &0.00  &0.00  &86.06  &0.00  &0.00  &112.41  \\
    SFT    &5.00  &8.34  &90.05  &1.04  &5.88  &89.25  &0.00  &0.00  &119.23  \\
    \textbf{TGIT(W)}
           & \textbf{7.50}{\scriptsize(+1.67)} & \textbf{8.56}{\scriptsize(-0.04)} & \textbf{92.67}{\scriptsize(-5.56)}
           & \textbf{1.56}{\scriptsize(+0.52)} & \textbf{4.03}{\scriptsize(+0.25)} & \textbf{88.48}{\scriptsize(-2.70)}
           & \textbf{0.00}{\scriptsize(+0.00)} & \textbf{0.00}{\scriptsize(+0.00)} & \textbf{120.41}{\scriptsize(-7.35)} \\
    \bottomrule
    \end{tabular}
    }
\end{table*}

\subsection{Results}
\label{sec:results}
\paragraph{The instruction gap is behavioral.}
On the same $203$ OpenFly Selected-Seen episodes and frozen navigator, SR drops from $31.03\%$ with Orig to $15.27\%$ with Weak and $11.33\%$ with Human.
Because the route, initial condition, controller, and evaluation remain fixed, this collapse isolates instruction source as the changed interface variable.
The lower Human result also shows that the gap is not merely an artifact of LM weakening.
Without Human-specific training, the same Weak-trained TGIT adapter raises Human-input SR from $11.33\%$ to $32.51\%$ ($2.87{\times}$), confirming that interface activation transfers beyond the Weak proxy.
Thus the deployment mismatch is behavioral, not only a difference in wording or length.

\paragraph{Trajectory grounding recovers the interface.}
Direct Weak execution reaches $15.27\%$ SR; prompt-only ICL improves modestly to $19.70\%$, and SFT target imitation recovers to $29.55\%$ but remains below Orig.
TGIT reaches $37.93\%$, gaining $22.66$ points over Weak, $8.38$ over SFT, and $6.90$ over Orig.
The controls identify the missing supervision: ICL rewrites without agent feedback, while SFT learns one executable target without comparing alternative commands under rollout outcomes.
TGIT instead executes and ranks candidates before updating only the translator.
This pattern is consistent with trajectory-grounded preference learning supplying information beyond prompting and target imitation; Section~\ref{sec:ablation} separately tests the hard-case and preference-design mechanisms.

\paragraph{Transfer reveals a backbone-bounded regime.}
Beyond the pool-construction environment, TGIT raises OpenFly SR from $14.11\%$ to $25.25\%$ on Unselected-Seen and from $4.95\%$ to $20.79\%$ on Unseen.
The relative activation also transfers across frozen agents: CityNav improves on Selected-Seen ($6.11\%\rightarrow10.00\%$), Unselected-Seen ($1.84\%\rightarrow9.20\%$), and Unseen ($0.00\%\rightarrow3.64\%$); AirVLN improves on Selected-Seen ($0.83\%\rightarrow7.50\%$) and Unselected-Seen ($0.00\%\rightarrow1.56\%$).
The OpenFly gains outside its $100$-case pool are inconsistent with a pool-memorization-only explanation, while the cross-agent gains show that the activation pattern is not specific to one navigator~\cite{wang2025rethinking,anderson2021sim}.
TGIT primarily expands success coverage rather than jointly optimizing every navigation metric.
Its OpenFly Selected-Seen SPL ($24.85\%$) remains below Orig and SFT, while the high Human SPL describes only the $11.33\%$ successful subset.
At the agent level, CityNav and AirVLN exhibit relative recovery but low absolute SR, and every AirVLN method remains at $0.00\%$ on Unseen.
This dual pattern---transferable relative activation but backbone-bounded absolute success---defines the claim: TGIT repairs recoverable interface failures but does not supply missing navigation skill.

\subsection{Ablation Study}
\label{sec:ablation}

\begin{table*}[t]
\centering
\caption{Multi-round ablation on data pool construction (OpenFly-Agent).
Results are \textbf{SR (\%) / SPL (\%) / NE (m)}.
\textbf{Selected}: original succeeds and weakened fails.
\textbf{Pool}: maintain 100 candidates and sample 20 per round.}
\label{tab:ablation_multi_round}
\resizebox{\textwidth}{!}{
\begin{tabular}{l|ccccc}
\toprule
\multirow{2}{*}{Variant}
& \multicolumn{5}{c}{\textbf{SR (\%) $\uparrow$ / SPL (\%) $\uparrow$ / NE (m) $\downarrow$}}  \\
\cmidrule(lr){2-6}
& Round 1 & Round 2 & Round 3 & Round 4 & Round 5 (Final) \\
\midrule
Random, Full
& 15.76 / 27.22 / 90.96
& 30.54 / 17.14 / 119.48
& 18.72 / 28.70 / 92.97
& 21.18 / 18.92 / 95.25
& 18.72 / 41.83 / 83.85 \\
Random, Pool
& 22.17 / 40.29 / 96.05
& 19.21 / 24.26 / 108.89
& 22.17 / 30.34 / 93.33
& 20.69 / 54.93 / 85.83
& 17.73 / 36.48 / 87.72 \\
\midrule
Selected, Full
& 31.03 / 15.97 / 119.20
& 33.50 / 21.68 / 110.30
& 28.08 / 23.35 / 101.37
& 31.03 / 31.79 / 103.51
& 29.06 / 24.27 / 105.22 \\
\textbf{Selected, Pool}
& 21.18 / 16.13 / 133.56
& 25.12 / 48.11 / 105.44
& 29.56 / 33.07 / 83.43
& 31.53 / 37.41 / 76.58
& \textbf{37.93 / 24.85 / 72.08} \\
\bottomrule
\end{tabular}
}
\end{table*}

\begin{figure*}[t]
\centering
\includegraphics[width=\textwidth]{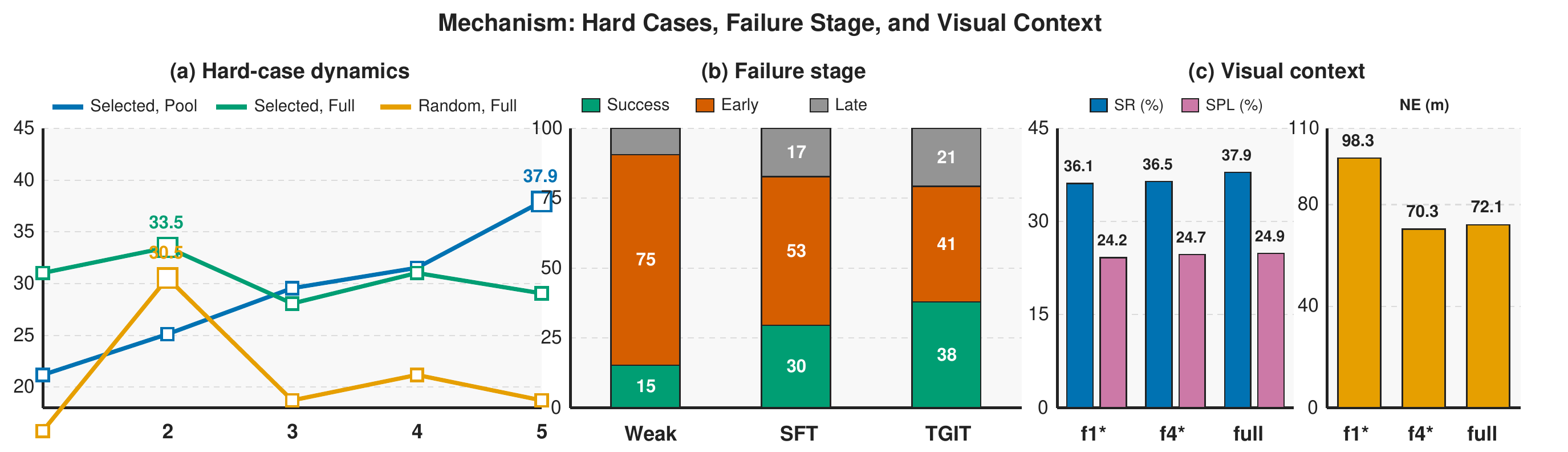}
\caption{\textbf{Hard cases, failure stage, and visual context.}
(a)~Pool curves (Table~\ref{tab:ablation_multi_round}); Selected-Pool Round~5 matches Table~\ref{tab:main_results}.
(b)~Failure stages with success fractions matched to Table~\ref{tab:main_results}; TGIT early/late proportions use the nearest Selected-Pool diagnostic after renormalizing success to $77/203$.
(c)~f1$^*$/f4$^*$ diagnostics; ``full'' is TGIT in Table~\ref{tab:main_results}.}
\label{fig:mechanism_panels}
\end{figure*}

We ablate TGIT on OpenFly-Agent (\emph{Airsim\_18}), which shows clear sensitivity to instruction style in Table~\ref{tab:main_results}.
The ablations isolate three factors behind that primary TGIT result: which episodes are supervised, how rollouts become preferences, and which translator backbone is used.

\paragraph{Hard-case selection and pool resampling.}
Table~\ref{tab:ablation_multi_round} compares random vs.\ hard-case selection and full-subset vs.\ pool resampling.
Random sampling is unstable (SR oscillates $15.76\rightarrow30.54\rightarrow18.72$).
Hard-case selection focuses on original-success / weak-failure episodes and yields stronger supervision.
Under hard-case selection, repeatedly training on a fixed full subset saturates, while resampling $K{=}20$ from a 100-case pool continues to improve through Round~5 and yields the TGIT result in Table~\ref{tab:main_results} ($37.93$ / $24.85$ / $72.08$).
Selection without resampling peaks early, while resampling without selection remains unstable; their combination supplies relevance and round-to-round diversity.

\paragraph{Mechanism: hard cases, failure stage, and visual context.}
Figure~\ref{fig:mechanism_panels} presents the three oral mechanism slots under a Table~\ref{tab:main_results}-aligned reading.
Selected-Pool improves through Round~5 to the TGIT result in Table~\ref{tab:main_results} ($37.93$), while Random-Full oscillates and Selected-Full saturates.
Weak failures are dominated by early collapse ($153/203$); SFT recovers to $29.55\%$ SR but leaves a large early-fail mass; TGIT expands success to $37.93\%$ and shrinks early failure.
Frame diagnostics approach the same TGIT numbers ($37.93$ / $24.85$ / $72.08$).
Across OpenFly/CityNav/AirVLN, relative activation with sharply different absolute levels separates interface repair from the backbone ceiling.

\paragraph{Reward ranking.}
Table~\ref{tab:reward_ablation} holds the Round~5 candidate set fixed and varies only how rollout metrics induce pairwise preferences.
It diagnoses ranking robustness among alternatives to the main Weighted form, not a substitute for the TGIT result in Table~\ref{tab:main_results} ($37.93\%$), which also resamples hard cases and chains adapters across rounds.
All success-prioritized forms recover well above Weak: Success-only reaches $35.47\%$, and lexicographic Success$\rightarrow$SPL$\rightarrow$NE reaches $36.14\%$, showing that the activation conclusion is not tied to the exact $0.5/0.3/0.2$ scalar weights.
We keep Weighted as the main-protocol reward because it supplies dense efficiency and distance signals when candidates are near-tied or unsuccessful; the diagnostic alternatives show that switching among success-first rankings does not reverse the preference-learning story.

\begin{table}[t]
\centering
\caption{\textbf{Reward Ranking} ($N{=}100$, Round~5).
Alternatives to the main Weighted form on a fixed candidate set.
Main TGIT SR remains $37.93\%$ (Table~\ref{tab:main_results}).}
\label{tab:reward_ablation}
\scriptsize
\setlength{\tabcolsep}{3pt}
\begin{tabular*}{\linewidth}{@{\extracolsep{\fill}}lcccc@{}}
\toprule
\textbf{Preference ranking} & \textbf{Valid} & \textbf{SR$\uparrow$} & \textbf{SPL$\uparrow$} & \textbf{NE$\downarrow$} \\
\midrule
Success only                 & 203/203 & 35.47 & 93.13 & 91.48 \\
Success$\rightarrow$SPL$\rightarrow$NE
                             & 202/203 & 36.14 & 95.00 & 84.60 \\
SPL$\rightarrow$NE           & 201/203 & 32.34 & 91.24 & 97.01 \\
\bottomrule
\end{tabular*}
\end{table}

\paragraph{Translator backbone and scale.}
Table~\ref{tab:ablation_model_size} compares translator backbones under the same diagnostic $N{=}100$, $M{=}4$ Round~5 protocol.
Qwen2.5-VL-7B is cleanest ($36.45\%$ SR, full coverage); 2B collapses valid outputs ($156/203$), so absolute SR becomes unreliable under a fixed $203$-episode denominator.
Because model families also differ (Qwen2.5/3/3.5), this is a practical backbone comparison rather than a monotonic parameter-scaling law.
It supports 7B as the clean default translator rather than proposing another end-to-end agent.

\begin{table}[t]
\centering
\caption{\textbf{Translator backbone/scale comparison} ($N{=}100$, $M{=}4$; diagnostic).
SR uses fixed denominator $203$; invalid outputs count as failures.
$^\dagger$NE is valid-only.
TGIT in Table~\ref{tab:main_results} remains the primary result.}
\label{tab:ablation_model_size}
\scriptsize
\setlength{\tabcolsep}{1.5pt}
\begin{tabular*}{\linewidth}{@{\extracolsep{\fill}}
>{\centering\arraybackslash}p{0.36\linewidth}
>{\centering\arraybackslash}p{0.17\linewidth}
>{\centering\arraybackslash}p{0.13\linewidth}
>{\centering\arraybackslash}p{0.13\linewidth}
>{\centering\arraybackslash}p{0.13\linewidth}
@{}}
\toprule
\textbf{Translator} & \textbf{Valid} & \textbf{SR} & \textbf{SPL} & \textbf{NE} \\
\midrule
Qwen2.5-VL-7B & 203/203 & \textbf{36.45} & \textbf{95.17} & 90.34 \\
Qwen3-VL-4B & 203/203 & 29.06 & 90.96 & 128.33 \\
Qwen3-VL-2B & 156/203 & 31.53 & 93.67 & 82.33$^\dagger$ \\
Qwen3.5-0.8B & 193/203 & 35.47 & 93.01 & 111.05$^\dagger$ \\
\bottomrule
\end{tabular*}
\end{table}

\paragraph{Discussion and claim boundaries.}
Hard-case selection isolates language as the changed variable when Orig succeeds and Weak fails; random/full ablations show that structure and resampling both matter.
Human SR $11.33\%$ shows the mismatch is not an LM-weakening artifact~\cite{dorbala2024can}, while Weak remains a stress protocol rather than a Human-distribution estimator; TGIT-on-Human is unmeasured.
TGIT's main gain is success coverage under the weighted reward ($+22.66$ SR over Weak; $+8.38$ over SFT), not lockstep SPL/NE improvement.
OpenFly shows the strongest activation; CityNav/AirVLN expose the backbone ceiling.

\section{Conclusion}

We studied an instruction-interface gap in aerial VLN: frozen agents can fail under short, underspecified inputs even when they possess the required navigation skill.
Human annotations establish the mismatch, while Weak provides a controlled stress protocol.
\textbf{TGIT} closes the interface loop through hard-case selection, trajectory-ranked preferences, and pool resampling while keeping the UAV agent frozen.
It raises OpenFly Selected-Seen SR from $15.27\%$ to $37.93\%$, exceeding Orig ($31.03\%$), ICL, and SFT.
CityNav/AirVLN expose the boundary: translation activates demonstrated skill but cannot strengthen the backbone.

\paragraph{Limitations and future work.}
TGIT currently performs one-shot translation before navigation and does not revise commands online as observations or execution states change.
A natural next step is latency-aware, real-time closed-loop translation that updates executable instructions during flight.
Moreover, the current agents remain limited in perception and control; another direction is to directly strengthen the navigation backbone itself.

\bibliography{reference}

\end{document}